\documentclass[11pt]{article}

\usepackage[a4paper,margin=1in]{geometry}
\usepackage{authblk}
\usepackage[round,authoryear]{natbib}
\setcitestyle{authoryear,round,semicolon}
\usepackage{graphicx}
\usepackage{booktabs}
\usepackage{multirow}
\usepackage{makecell}
\usepackage{amsmath,amsfonts,amssymb}
\usepackage{tabularx}
\usepackage{xcolor}
\usepackage[
    colorlinks=true,
    linkcolor=blue,
    citecolor=blue,
    urlcolor=blue
]{hyperref}

\newcommand{\figdir}{figures}

\title{A Seismic Foundation Model for Dense Feature Extraction}
\author[1,2,3]{Zhixiang Guo}
\author[1,2,3]{Xinming Wu\thanks{Corresponding author: xinmwu@ustc.edu.cn}}
\author[1,2,3]{Yimin Dou}
\author[4]{Hui Gao}
\author[1,2,3]{Siyao Zhao}
\author[1,2,3]{He Huang}
\author[1,2,3]{Wei Liu}
\affil[1]{Laboratory of Seismology and Physics of the Earth's Interior, School of Earth and Space Sciences, University of Science and Technology of China, Hefei 230026, China.}
\affil[2]{State Key Laboratory of Precision Geodesy, University of Science and Technology of China, Hefei 230026, China.}
\affil[3]{Mengcheng National Geophysical Observatory, University of Science and Technology of China, Hefei 230026, China.}
\affil[4]{School of Artificial Intelligence, Anhui University of Science and Technology, Hefei 231131, China}
\date{}

\begin{document}
\maketitle

\begin{abstract}
Deep-learning-based seismic interpretation is constrained by 
scarce high-quality annotations, and most existing models are trained separately 
for individual tasks. We present SeisDFE, a 7B-parameter seismic foundation model 
for dense feature extraction, pretrained on 5.34 million 2D images derived from more 
than 500 seismic volumes. 
A three-stage pretraining strategy — discriminative self-supervised learning, 
Gram feature anchoring, and multi-resolution adaptation — improves feature 
discriminability while preserving spatial coherence. 
With its encoder frozen, SeisDFE supports seismic facies classification, 
salt detection, fault detection, and relative geologic time estimation. 
It outperforms existing seismic foundation models on all four tasks and a 
natural-image-pretrained model on all but fault detection, with the largest 
gains under scarce supervision: using only two annotated sections of the F3 dataset, 
it improves whole-volume mIoU by 12–19 percentage points over the baselines. 
The learned features remain effective with task-specific decoders and extend 
to synthetic-to-field transfer and sparse 3D geobody interpretation. 
These results show that large-scale seismic pretraining organizes geological 
information into dense representations that are directly reusable across 
interpretation tasks.
\end{abstract}


\section{Introduction}
\label{sec:introduction}

Seismic interpretation is fundamental to characterizing subsurface structure,
stratigraphic architecture, and geologic bodies from reflection seismic data.
It provides essential information for geological understanding and for decisions
throughout seismic exploration and subsurface development
\citep{geng2020automated,yu2021deep,mousavi2022deep,khosro2024machine}.
Deep learning has substantially advanced the automation of seismic interpretation,
with specialized models developed for fault detection
\citep{xiong2018seismic,wu2019faultseg3d,gao2022automatic,dou2024faultssl},
seismic facies classification
\citep{liu2020seismic,zhang2021automatic,kaur2023deep},
geobody delineation
\citep{gao2021channelseg3d,yan20253d},
and horizon interpretation
\citep{geng2020deep,roncoroni2022efficient}.
These advances still depend heavily on scarce field interpretations
or large-scale synthetic training data \citep{wu2023sensing}. High-quality field labels are costly and
time-consuming to obtain, while models trained mainly on synthetic data may suffer
from distribution shifts when applied to field surveys. These limitations become
particularly severe when only a few annotated sections are available for a new
interpretation problem.

Despite their different targets, many seismic interpretation tasks rely on common
patterns in reflection data. Seismic facies are characterized by recurring variations
in reflection amplitude, texture, and geometry that reflect different depositional
and stratigraphic settings \citep{alaudah2019machine,di2019developing,zhang2021automatic}. 
Faults appear as local disruptions or offsets of otherwise
continuous reflection events \citep{xiong2018seismic,wu2019faultseg3d}. 
Geobodies such as salt structures and channels produce
distinct changes in reflection amplitude, morphology, and continuity because their
properties differ from those of the surrounding formations \citep{gao2021channelseg3d}. 
Although their prediction targets differ, these tasks all draw on recurring reflection 
patterns at different spatial scales \citep{harsuko2022storseismic,wu2019multitask}.
Most task-specific deep-learning workflows learn representations separately for
individual targets and datasets. As a result, the same structural information 
is relearned for each task from limited annotations. This redundancy motivates 
a shared seismic representation that can be reused across interpretation tasks.

Foundation models offer a route toward reusable seismic representations.
Pretrained vision models can transfer useful features to geophysical data
\citep{guo2025cross}, while recent seismic foundation models have demonstrated the
feasibility of learning from large collections of unlabeled seismic data
\citep{sheng2025seismic,
cheng2025generativefoundationmodelallinone,
dong2025lightweightedfoundationmodelseismic,
gao2026foundation,
ordonez2026ncs,
archibong2025seiscode}.
Their downstream applications commonly rely on full fine-tuning or
parameter-efficient encoder adaptation
\citep{chai2020using,hu2021lora}.
Masked and reconstruction-based objectives \citep{he2022masked,bao2021beit}, 
adopted by several seismic foundation models, learn contextual dependencies by 
recovering missing content but do not explicitly encourage geological patterns to 
be separable or patch relations to remain spatially stable.
Both properties are especially important in extreme few-shot settings, where downstream supervision
is insufficient to substantially reorganize the pretrained feature space \citep{zhou2023foundation}.

This raises a central question for seismic foundation models: can large-scale
unlabeled seismic data be organized during pretraining into a dense representation
that is discriminative, spatially coherent, and directly reusable across interpretation
tasks? We hypothesize that reusable dense representations require both feature
discrimination and spatial coherence. Discriminative self-supervised learning can
promote separation in feature space
\citep{caron2021emerging,zhou2021ibot,oquab2023dinov2},
while constraining feature relations keeps the spatial associations among geologically related features stable
\citep{bardes2022vicregl,wen2025semantic,simeoni2025dinov3}.
We therefore investigate whether discriminative seismic pretraining combined with
explicit preservation of dense feature relations can produce representations that
remain directly reusable across interpretation tasks.

\begin{figure}[t]
\centering
\includegraphics[width=0.99\textwidth]{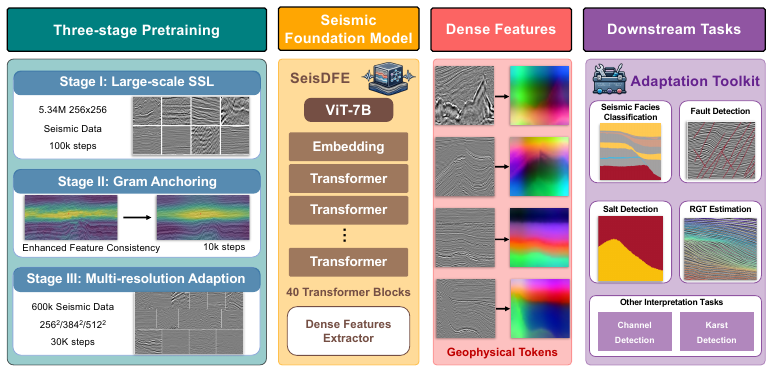}
\caption{Overall framework of SeisDFE. The left panel shows the three-stage pretraining pipeline, consisting of large-scale self-supervised learning, Gram feature anchoring, and multi-resolution adaptation. The middle panel shows the frozen ViT-7B seismic foundation model and its patch-level dense features. The right panel shows the adaptation of task-specific decoders for seismic facies classification, fault detection, salt detection, RGT estimation, and other interpretation tasks.}
\label{fig1}
\end{figure}

Here we test this hypothesis with SeisDFE, a 7B-parameter seismic foundation model
pretrained on 5.34 million unlabeled 2D images derived from more than 500 seismic
volumes. 
The overall framework is illustrated in Figure~\ref{fig1}. SeisDFE adopts a
progressive three-stage pretraining strategy. Stage~I performs large-scale
discriminative self-supervised learning on unlabeled seismic data. During Stage~I,
feature-similarity analysis reveals that prolonged optimization can coincide with a
loss of local spatial coherence. This observation motivates Stage~II, which introduces
Gram feature anchoring to constrain dense feature relations during continued
pretraining. Stage~III further introduces multi-resolution adaptation to improve the
consistency of the learned representation across input scales. The resulting encoder
serves as a shared dense feature extractor that can be coupled with lightweight
task-specific heads or decoders for different seismic interpretation problems.

We evaluate whether the resulting representation can be reused without repeated
encoder optimization. With the encoder frozen, SeisDFE supports seismic facies
classification, salt detection, fault detection, and relative geologic time estimation
across different levels of supervision. The largest gains occur when annotations are
scarce, suggesting that substantial task-relevant geological information is already
accessible from the pretrained feature space. The learned representation also remains
effective with stronger task-specific decoders and can be extended to field-data and
sparse three-dimensional applications. These results support a role for seismic
foundation models beyond task-specific initialization, as a shared feature basis for
data-efficient seismic interpretation.

The main contributions of this study are as follows:

\begin{itemize}

\item We construct a large-scale seismic pretraining dataset
comprising 5.34 million unlabeled seismic images derived from
more than 500 seismic volumes. The dataset covers diverse
geological structures, sedimentary basins, and acquisition
conditions, providing a broad basis for
learning transferable seismic representations.

\item We develop SeisDFE, a 7B-parameter seismic foundation
model that adapts large-scale discriminative self-supervised
learning, Gram feature anchoring, and multi-resolution
training to seismic data. The progressive pretraining strategy
promotes feature discriminability and spatial coherence,
enabling the extraction of reusable dense seismic features.
We further distill SeisDFE into a computationally efficient
variant, SeisDFE-Large.

\item We systematically evaluate the cross-task reusability
and practical applicability of SeisDFE. Through frozen-encoder
evaluation, the learned representations demonstrate strong
performance across seismic facies classification, salt
detection, fault detection, and RGT estimation, particularly
under limited supervision. We further extend the evaluation
to synthetic-to-field transfer and sparse 3D geobody
interpretation, showing that pretrained seismic features support 
field-data applications and volumetric prediction.

\end{itemize}


\section{Method}
\label{sec:method}

\subsection{Pretraining Data}

The transferability of a seismic foundation model depends strongly on the diversity
of structures and acquisition conditions represented in its pretraining 
data \citep{schafer2024overcoming}.
We included
the 3D seismic data used by SFM \citep{sheng2025seismic} and further collected
publicly available migrated 3D seismic volumes from the United States Geological
Survey (USGS), the Society of Exploration
Geophysicists (SEG), New Zealand
Petroleum and Minerals (NZPAM), the
Netherlands Oil and Gas Portal (NLOG), and
other public data platforms. After format unification, duplicate removal, and quality
control, the resulting dataset contains more than 500 3D seismic volumes covering
over 200,000~$\mathrm{km}^{2}$. These data span different sedimentary basins,
acquisition vintages, processing workflows, signal-to-noise ratios, and spatial
sampling conditions, and contain diverse geological structures such as faults, folds,
unconformities, and geobodies.

As SeisDFE operates on 2D inputs, each 3D seismic
volume was sampled along the inline and crossline
directions. Sections containing more than 1\% zero-valued
samples were excluded to remove data dominated by missing samples or dead
traces. A minimum spatial separation was imposed between neighboring sampling
positions to reduce the redundancy caused by adjacent seismic sections.

The retained sections were center-cropped to $512\times512$ pixels.
For large-scale pretraining, these sections were further divided into overlapping
$256\times256$ patches. This procedure produced 5.34 million
$256\times256$ seismic images. A separate set of 600,000
$512\times512$ images was retained for the later pretraining stages.
All field seismic surveys used for downstream evaluation were distinct
from those included in the pretraining corpus.

Amplitude distributions vary across surveys because of differences in acquisition
and processing. To reduce the influence of extreme amplitudes, each image was
clipped to
$[\mu-3\sigma,\mu+3\sigma]$,
where $\mu$ and $\sigma$ denote the mean and standard deviation of the image,
respectively. Mean-variance standardization was then applied, followed by
min-max normalization to map the amplitudes to $[0,1]$.

\subsection{Three-stage Progressive Pretraining}

\begin{figure}[t]
\centering
\includegraphics[width=0.99\textwidth]{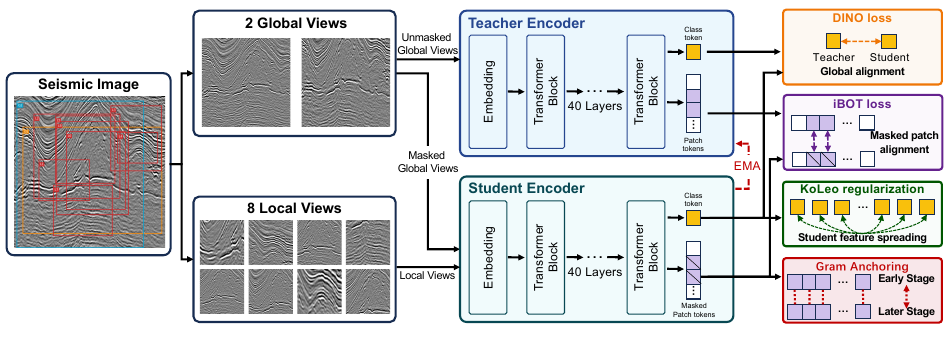}
\caption{The discriminative self-supervised pretraining framework of SeisDFE. Each
seismic image yields two global views and eight local views, which are fed to the
teacher and student encoders respectively, with the teacher parameters updated by an
exponential moving average of the student parameters. The training objective
comprises the DINO loss, the iBOT loss, the KoLeo regularizer, and the Gram anchoring
loss.}
\label{contrastive}
\end{figure}

SeisDFE uses three stages of progressive pretraining to adapt visual
representations to seismic data and organize spatially resolved geological
information. Stage~I develops discriminative seismic features through
large-scale self-supervised learning. Stage~II stabilizes the relations among
patch features identified during Stage~I, and Stage~III adapts the representation
to different input fields of view. The teacher--student pretraining framework
and its objectives are illustrated in Figure~\ref{contrastive}.

\subsubsection{Stage I: Discriminative Self-supervised Pretraining}

SeisDFE uses a ViT-7B encoder with 40 Transformer blocks. The model is
initialized from DINOv3 ViT-7B \citep{simeoni2025dinov3}. This initialization
avoids training a 7B-parameter encoder from random weights and provides a
general visual representation that can be further adapted to seismic data
\citep{guo2025cross}.

The original DINOv3 encoder accepts three-channel natural images, whereas seismic
sections contain a single amplitude channel. We therefore modify the
patch-embedding layer to accept single-channel input. The new embedding weights
are initialized by averaging the original RGB weights along the input-channel
dimension. The remaining encoder parameters are initialized directly from the
pretrained DINOv3 model.

For an input seismic section, the encoder produces a sequence of patch tokens.
These spatially resolved tokens are used as dense seismic features in downstream
tasks. Features from intermediate Transformer blocks can also be extracted when
multi-level representations are required by a downstream decoder.

Stage~I adapts the pretrained visual representation to the seismic domain using
discriminative self-supervised learning. Each seismic image is transformed into
two global views and multiple local views. The views are processed by a
teacher-student network following the DINO and iBOT framework
\citep{caron2021emerging,zhou2021ibot,oquab2023dinov2}. 
The teacher is updated
from the student using an exponential moving average; backpropagation
updates the student parameters.

The global views provide section-scale structural context, whereas the local
views emphasize smaller reflection patterns (Figure~\ref{contrastive}). 
Masking is additionally applied to
the student inputs so that local representations must be inferred from their
surrounding context. Training combines image-level DINO self-distillation,
patch-level iBOT prediction, and KoLeo feature-spreading regularization.

For the image-level objective, let $x_g$ denote a global view processed by the
teacher and $x_v$ another global or local view processed by the student.
The DINO loss is written as
\begin{equation}
\mathcal{L}_{\mathrm{DINO}}
=
\sum_{g}
\sum_{v\neq g}
D_{\mathrm{KL}}
\left(
q_t(x_g)\,\|\,p_s(x_v)
\right),
\end{equation}
where $q_t$ and $p_s$ denote the teacher and student output distributions,
respectively. The teacher output is treated as a fixed target during
backpropagation.

Dense seismic interpretation also requires information at individual spatial
locations. The iBOT objective therefore constrains the representation at masked
patch positions:
\begin{equation}
\mathcal{L}_{\mathrm{iBOT}}
=
\frac{1}{|\mathcal{M}|}
\sum_{i\in\mathcal{M}}
D_{\mathrm{KL}}
\left(
q_t^{\,i}\,\|\,p_s^{\,i}
\right),
\end{equation}
where $\mathcal{M}$ is the set of masked patch positions, and
$q_t^{\,i}$ and $p_s^{\,i}$ are the corresponding teacher and student
distributions.

The Stage~I objective is
\begin{equation}
\mathcal{L}_{\mathrm{Stage\,I}}
=
\frac{8}{9}\mathcal{L}_{\mathrm{DINO\text{-}local}}
+
\frac{1}{9}\mathcal{L}_{\mathrm{DINO\text{-}global}}
+
\mathcal{L}_{\mathrm{iBOT}}
+
0.2\mathcal{L}_{\mathrm{KoLeo}}.
\end{equation}
Here, $\mathcal{L}_{\mathrm{DINO\text{-}local}}$ denotes alignment between
teacher global views and student local views, whereas
$\mathcal{L}_{\mathrm{DINO\text{-}global}}$ denotes alignment between the
global views. KoLeo regularization is applied to the student representation to
discourage excessive concentration in feature space.
Stage~I is trained for 100k steps using the 5.34 million
$256\times256$ seismic images.

To monitor dense feature relations during Stage~I, we examine the
evolution of patch-to-patch similarity. A seismic section from the F3 dataset
\citep{alaudah2019machine} is passed through checkpoints saved at 50k, 80k,
and 100k training steps. A reference patch is selected, and its cosine similarity
with every other patch feature is calculated. The similarities are then mapped
back to their spatial positions to visualize the spatial extent of related features.

\begin{figure}[t]
\centering
\includegraphics[width=0.99\textwidth]{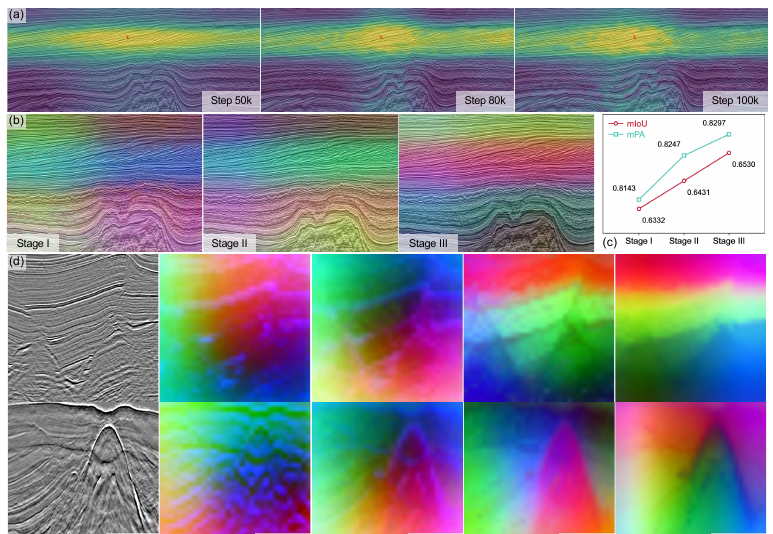}
\caption{Evolution and organization of dense seismic representations.
(a) Patch-feature similarity during Stage~I at 50k, 80k, and 100k training steps.
(b) PCA visualization of dense features after the three pretraining stages.
(c) Frozen linear-probing performance of the three stages on SEAM facies classification.
(d) PCA visualization of features extracted from different depths of the final SeisDFE encoder.}
\label{fig2}
\end{figure}

At 50k training steps, patches associated with the same continuous
stratal unit show concentrated and spatially coherent similarity responses
(Figure~\ref{fig2}a). As training proceeds to 80k and 100k steps, these
responses become increasingly fragmented, with high similarities appearing
at structurally unrelated locations. Thus, continued self-supervised
optimization is accompanied by a gradual degradation of local spatial
relations in the dense feature space. Similar behavior has been reported
during prolonged discriminative pretraining of large vision models
\citep{simeoni2025dinov3}.

\subsubsection{Stage II: Gram Feature Anchoring}

Continued Stage~I training changes the relations among dense patch features even
when the self-supervised objective continues to decrease. Stage~II therefore
introduces a Gram-based relational constraint to stabilize these spatial
relationships.

The student and self-supervised teacher are initialized from the final Stage~I
model. The self-supervised teacher continues to follow the student using an
exponential moving average with a momentum of 0.999 and provides targets for the
DINO and iBOT objectives. A second teacher is introduced only for Gram
anchoring. This teacher is frozen throughout Stage~II.

The EMA teacher saved at 50k steps of Stage~I is used as the Gram teacher.
At this checkpoint, the representation had already acquired seismic-domain
structure while retaining the spatial coherence observed in the feature-similarity
analysis, whereas later checkpoints showed the fragmentation described above.

Stage~II uses the 600,000 $512\times512$ seismic images. The student retains
the multi-view strategy used in Stage~I. Global and local views are resampled to
$256\times256$ and $112\times112$, respectively. The Gram teacher receives
the same spatial crops at twice the student resolution. Its feature maps are then
downsampled to the student token grid before the relational loss is evaluated.
This provides a higher-resolution relational target while allowing the teacher and
student Gram matrices to be compared on the same spatial grid
\citep{wysoczanska2024clip}. Rotary positional embeddings allow the encoder to
process different input sizes without resolution-specific positional parameters
\citep{su2024roformer}.

Let the normalized student and Gram-teacher patch features be
\begin{equation}
\mathbf{F}\in\mathbb{R}^{N\times d},
\qquad
\mathbf{F}_{a}\in\mathbb{R}^{N\times d},
\end{equation}
where $N$ is the number of spatial tokens and $d$ is the feature dimension.
Their Gram matrices are
\begin{equation}
\mathbf{G}
=
\mathbf{F}\mathbf{F}^{\mathrm T},
\qquad
\mathbf{G}_{a}
=
\mathbf{F}_{a}\mathbf{F}_{a}^{\mathrm T}.
\end{equation}

Because the patch features are normalized, each Gram-matrix element corresponds
to the cosine similarity between two spatial positions. The Gram loss is defined
as
\begin{equation}
\mathcal{L}_{\mathrm{Gram}}
=
\frac{1}{N^{2}}
\left\|
\mathbf{G}-\mathbf{G}_{a}
\right\|_{\mathrm F}^{2}.
\end{equation}

The Stage~II objective is
\begin{equation}
\mathcal{L}_{\mathrm{Stage\,II}}
=
\mathcal{L}_{\mathrm{SSL}}
+
\lambda_{\mathrm{Gram}}
\mathcal{L}_{\mathrm{Gram}},
\end{equation}
where $\mathcal{L}_{\mathrm{SSL}}$ contains the DINO, iBOT, and KoLeo terms
retained from Stage~I, and $\lambda_{\mathrm{Gram}}$ controls the strength of
relational anchoring. Stage~II is trained for 10k steps.

\subsubsection{Stage III: Multi-resolution Adaptation}

Stage~III adapts the learned representation to changes in input field of view and
token-grid resolution. It is initialized from the Stage~II model and uses the same
600,000 $512\times512$ seismic images.

At each training iteration, the student receives either the full
$512\times512$ image or a randomly cropped $384\times384$ or
$256\times256$ region. All input sizes are derived from the same seismic
images and therefore have the same physical sampling interval. 
Varying the crop size therefore changes the field of view and token-grid 
size without resampling traces or time samples.

The self-supervised objectives and Gram constraint are retained during this stage.
This allows the model to adapt to different input sizes while preserving the
feature relations learned during Stage~II. Stage~III is trained for a further
30k steps.

\subsection{Model Distillation}

A smaller SeisDFE-Large model is obtained by distilling the final SeisDFE-7B
representation. The student is initialized from DINOv3-Large, while the final
SeisDFE-7B serves as the teacher.

Distillation uses the same 5.34 million unlabeled seismic images employed in
Stage~I. The student is trained for 100k steps to match both the image-level DINO
outputs and the patch-level iBOT outputs produced by SeisDFE-7B. This transfers
both global and spatially resolved information from the 7B teacher to the smaller
encoder. SeisDFE-Large is used in the downstream experiments to evaluate how
much of the learned representation is retained after model compression.

\subsection{Implementation Details}

The three SeisDFE pretraining stages were performed on NVIDIA H20 GPUs with
96~GB of memory. Stage~I, Stage~II, and Stage~III were trained for 100k,
10k, and 30k steps, respectively. Their global batch sizes were 64, 24, and 56.
The DINO-local loss weights were set to 1.0, 0.7, and 0.5 across the three
stages, respectively. The Gram loss weights were set to 1.0 in Stage~II
and 1.5 in Stage~III. The corresponding computational costs were
approximately 428, 39, and 435 GPU-hours.

\subsection{Dense Feature Analysis}

We characterize the learned dense representation through spatial
visualization and matched frozen-encoder probing. The Stage~I checkpoint
analysis above establishes the motivation for preserving patch-feature
relations. The following analyses examine feature organization across the
three stages and across Transformer depths.

We use principal component analysis (PCA) to visualize the dense feature
maps produced after Stages~I-III. Patch features are projected onto the first
three principal components, which are mapped to the three image channels.
The same visualization procedure is also applied to features extracted from
different depths of the final SeisDFE encoder.

We also perform frozen linear probing on the SEAM seismic facies dataset \citep{seam}
to compare the three pretraining stages. The encoder is frozen for all checkpoints,
and the same linear prediction head, training samples, and optimization protocol
are used. Performance is evaluated using mean intersection over union (mIoU) and
mean pixel accuracy (mPA).

After Stage~II, which incorporates Gram feature anchoring, the dense
representations appear more coherent along reflection events and within
geobodies (Figure~\ref{fig2}b). Major structural boundaries also remain
more clearly separated. Stage~III retains a similar organization across
different input sizes, suggesting that the learned feature relations remain
more stable when the field of view and token-grid size change.

We next test whether these representational changes are accompanied by improved
downstream accessibility. Using the same frozen linear-probing protocol on the
SEAM facies dataset \citep{seam}, Stage~II increases mIoU from 0.6332 to
0.6431 and mPA from 0.8143 to 0.8247. Stage~III further reaches 0.6530 mIoU
and 0.8297 mPA (Figure~\ref{fig2}c). Because the downstream protocol is
identical across the three checkpoints, these improvements are consistent with
progressively greater accessibility of task-relevant information in the learned
representation.

The final SeisDFE representation also shows a qualitative hierarchy across
network depth (Figure~\ref{fig2}d). Shallow-layer features emphasize local
amplitude variations, reflection textures, and structural edges. Intermediate
layers show broader organization along reflection systems and stratigraphic
variations, whereas deeper layers form larger and more coherent feature regions
while retaining major structural boundaries. These patterns emerge without
interpretation labels and suggest a progression from local reflection attributes
toward broader geological organization.

\section{Evaluation and Applications}
\label{sec:results}

We evaluate the reuse of SeisDFE representations across increasing levels of
downstream supervision and spatial context. Extreme few-shot linear probing
tests the direct accessibility of task information from frozen features.
Task-specific decoders examine the representation under larger annotated
training sets. We then evaluate synthetic-to-field transfer and sparse
three-dimensional interpretation using the same pretrained encoder.
The experimental data, model configurations, and supervision protocols are
described together with their corresponding results below.

\subsection{Extreme Few-shot Linear Probing}

Extreme few-shot experiments are designed to measure how directly downstream
information can be accessed from each pretrained representation. All encoder
parameters are frozen, and only a linear prediction layer is trained at each
spatial position.

We compare SeisDFE and SeisDFE-Large with DINOv3 ViT-7B
\citep{simeoni2025dinov3}, SFM-Large-512
\citep{sheng2025seismic}, and NCS-2D \citep{ordonez2026ncs}.
The same downstream head and training samples are used for all encoders.
Inputs are normalized according to the preprocessing used by each model during
its original pretraining.

Salt detection uses the TGS Salt Identification Challenge dataset
\citep{tgs}. The dataset contains 4,000 annotated seismic images of
size $224\times224$. We use 3,000 images as the training pool and 1,000 for
validation. Few-shot training subsets contain 5, 10, 20, 50, or 100 images.

\begin{figure}[t]
\centering
\includegraphics[width=0.99\textwidth]{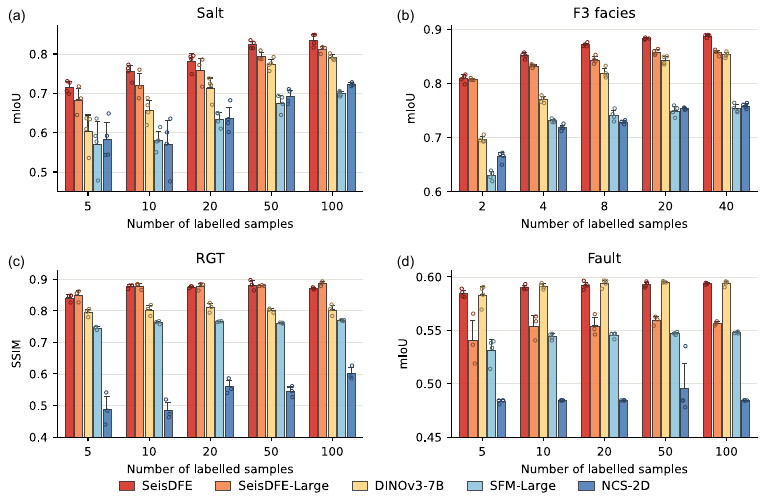}
\caption{Extreme few-shot linear probing with frozen encoders.
(a) Salt detection, (b) F3 facies classification,
(c) RGT estimation, and (d) fault detection.
Only a linear prediction head is trained for each task.}
\label{fig3}
\end{figure}

Seismic facies classification uses the F3 dataset
\citep{alaudah2019machine}. The seismic volume has size
$255\times701\times401$ in the time, crossline, and inline directions.
Each inline section has an original size of $255\times701$ and is resized to
$256\times704$ using bilinear interpolation. The labels are resized using
nearest-neighbor interpolation. We use 2, 4, 8, 20, or 40 annotated inline
sections for training. Inference is then performed over the complete 3D volume.

RGT estimation uses the geologically constrained synthetic stratigraphic dataset
of \citet{gao2025geologically}. It contains 4,000 seismic images of size
$512\times512$, of which 3,200 are used as the training pool and 800 for
validation. Few-shot subsets contain 5, 10, 20, 50, or 100 training samples.

Fault detection uses the manually interpreted field seismic dataset of
\citet{guo2025cross}. The dataset contains 1,350 images of size
$896\times896$, with 1,081 used as the training pool and 269 for validation.
The same few-shot sample sizes used for salt and RGT estimation are evaluated.

The three segmentation tasks are optimized using the weighted Dice loss adopted
by \citet{guo2025cross}. RGT estimation uses mean squared error. Salt detection,
facies classification, and fault detection are evaluated using mIoU, whereas RGT
estimation is evaluated using structural similarity index measure (SSIM).
Each few-shot setting is repeated four times using different random
selections of the training samples, and the distributions of the
resulting scores are reported.

SeisDFE shows the clearest advantage for salt detection and seismic facies
classification (Figure~\ref{fig3}a, b). With only five annotated salt images, SeisDFE reaches an mIoU of
0.72, compared with 0.60 for DINOv3-7B, 0.58 for NCS-2D, and 0.57 for
SFM-Large. The advantage persists as more annotations are introduced. An even
stronger result is observed on the F3 facies dataset. Using only two annotated
sections, SeisDFE reaches a whole-volume mIoU of 0.82. DINOv3-7B, NCS-2D, and
SFM-Large reach 0.70, 0.66, and 0.63, respectively. 
Because supervision is provided on only a few 2D sections while inference covers
the full 3D volume, this result is consistent with stratal and facies information
being readily accessible from the pretrained feature space.

RGT estimation depends even less on the number of labels (Figure~\ref{fig3}c).
SeisDFE reaches an SSIM of approximately 0.84 with only five training samples
and improves only moderately as more labels are added. 
SeisDFE-Large performs on par with SeisDFE at all sample sizes.
This behavior suggests that much of the reflection continuity and stratigraphic
information required for RGT estimation is already accessible in the pretrained
representation before downstream training.

Fault detection follows a different pattern (Figure~\ref{fig3}d). 
SeisDFE and DINOv3-7B perform comparably at all sample sizes, and both 
clearly outperform SFM-Large and NCS-2D, whereas SeisDFE-Large declines 
moderately after distillation. Faults are sparse, narrow discontinuities 
that are difficult to resolve with a single linear mapping on a coarse 
token grid. Seismic-domain pretraining therefore adds less for this localized 
target than for regional facies, geobodies, and continuous stratigraphy, 
and fault detection relies more on spatial resolution and decoder capacity, 
which we examine next.

SeisDFE-Large retains much of the few-shot capability of the 7B model,
particularly for facies classification and RGT estimation, and the results remain
stable across different training subsets. These experiments isolate how readily task-relevant information can be accessed
from the foundation models while keeping their encoders frozen. We next
examine whether this advantage persists with more supervision and stronger
task-specific decoders.

\subsection{Evaluation with Task-specific Decoders}

We further evaluate the pretrained representations using larger training sets and
task-specific dense prediction decoders. SeisDFE, SeisDFE-Large, and
DINOv3-7B remain frozen throughout these experiments.

For seismic facies classification, we use the SEAM split adopted by \citet{sheng2025seismic}, which contains 94 training and 23 validation images
of size $768\times768$. A PUP decoder \citep{zheng2021rethinking} combines
multi-level encoder features and progressively restores spatial resolution.

Salt detection uses the full TGS training set described above. A linear prediction
head is retained to test whether the feature separability observed in the few-shot
setting persists when more annotations are available.

Fault detection uses the full field dataset described above. An MLA decoder
\citep{zheng2021rethinking} combines features extracted from different encoder
depths to recover narrow structural discontinuities at higher spatial resolution.

For SFM-Large and NCS-2D, both frozen and LoRA-adapted configurations are
evaluated, and the configuration with the better validation performance is reported.
This protocol favors the baselines, because SeisDFE is evaluated only with a frozen encoder.
Under this protocol, SFM-Large uses LoRA for all three tasks. NCS-2D remains
frozen for seismic facies classification and uses LoRA for salt and fault detection.








\begin{table}[t]
\centering
\begin{minipage}{0.7\textwidth}
\centering
\caption{Performance of different pretrained encoders on three downstream seismic
interpretation tasks. Bold indicates the best result.}
\label{tab:downstream}
\setlength{\tabcolsep}{6pt}
\renewcommand{\arraystretch}{1.12}
\resizebox{\linewidth}{!}{
\begin{tabular}{lllccc}
\toprule
Task & Decoder & Encoder & \makecell{Encoder\\setting} & mIoU & mPA \\
\midrule
\multirow{5}{*}{\makecell[c]{SEAM facies\\classification}}
& \multirow{5}{*}{PUP}
& DINOv3-7B       & Frozen & 0.8374 & 0.9221 \\
& & SFM-Large     & LoRA   & 0.8243 & 0.8994 \\
& & NCS-2D        & Frozen & 0.8415 & 0.9121 \\
& & SeisDFE-Large & Frozen & 0.8768 & 0.9289 \\
& & SeisDFE       & Frozen & \textbf{0.8845} & \textbf{0.9324} \\
\midrule
\multirow{5}{*}{\makecell[c]{TGS salt\\detection}}
& \multirow{5}{*}{Linear}
& DINOv3-7B       & Frozen & 0.8116 & 0.8815 \\
& & SFM-Large     & LoRA   & 0.8044 & 0.8773 \\
& & NCS-2D        & LoRA   & 0.8046 & 0.8767 \\
& & SeisDFE-Large & Frozen & 0.8565 & 0.9149 \\
& & SeisDFE       & Frozen & \textbf{0.8819} & \textbf{0.9297} \\
\midrule
\multirow{5}{*}{\makecell[c]{Fault\\detection}}
& \multirow{5}{*}{MLA}
& DINOv3-7B       & Frozen & \textbf{0.7564} & \textbf{0.8023} \\
& & SFM-Large     & LoRA   & 0.6361 & 0.7062 \\
& & NCS-2D        & LoRA   & 0.6468 & 0.7165 \\
& & SeisDFE-Large & Frozen & 0.7090 & 0.7691 \\
& & SeisDFE       & Frozen & 0.7374 & 0.7867 \\
\bottomrule
\end{tabular}}
\end{minipage}
\end{table}

\begin{figure}[t]
\centering
\includegraphics[width=0.99\textwidth]{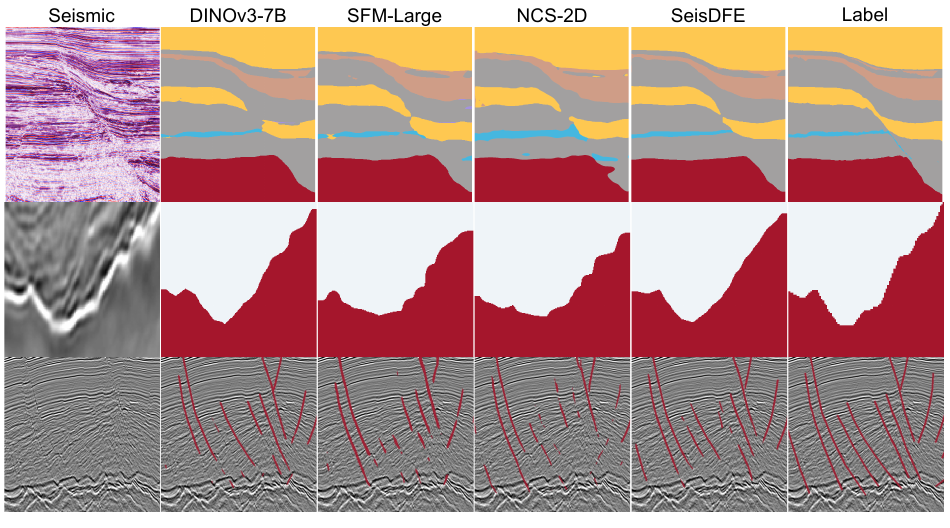}
\caption{Qualitative comparisons for downstream seismic interpretation.
The rows show seismic facies classification, salt detection, and fault detection.
The columns show the seismic input, predictions from DINOv3-7B, SFM-Large,
NCS-2D, and SeisDFE, and the reference interpretation.}
\label{fig:downstream_qualitative}
\end{figure}

The quantitative comparison is summarized in Table~\ref{tab:downstream},
and representative predictions for the three interpretation tasks are
shown in Figure~\ref{fig:downstream_qualitative}.

SeisDFE achieves the best performance on seismic facies classification and salt
detection. On the SEAM facies dataset, it reaches an mIoU of 0.8845 and an mPA
of 0.9324, compared with 0.8374 and 0.9221 for DINOv3-7B. SeisDFE-Large
reaches an mIoU of 0.8768, retaining most of the performance of the 7B model.
The advantage is also clear for salt detection. SeisDFE reaches an mIoU of
0.8819, compared with 0.8116 for DINOv3-7B, 0.8044 for SFM-Large, and
0.8046 for NCS-2D. 
These results show that the advantage observed in the few-shot setting persists
when more supervision is available.

Fault detection again behaves differently. With the MLA decoder, SeisDFE 
reaches an mIoU of 0.7374, exceeding SFM-Large (0.6361) and NCS-2D (0.6468) 
by about 9–10 percentage points while remaining slightly below 
DINOv3-7B (0.7564). With multi-level fusion and the full training set, 
fault mIoU rises from below 0.60 in the few-shot setting to above 0.73, 
indicating that narrow discontinuities benefit more from multi-level 
fusion and spatial decoding than regional targets do.

Across these downstream settings, the frozen SeisDFE representation remains
effective for facies and salt interpretation, while the advantage is smaller for
fault detection. SeisDFE-Large retains much of the performance of the 7B
model for facies and salt, but shows a larger reduction on faults. 
We further examine whether the pretrained representation can also support transfer from
synthetic training data to field seismic surveys.

\subsection{Synthetic-to-field Transfer}
\label{sec:synthetic_to_field}

We construct large-scale synthetic training data for fault
detection, karst detection, and RGT estimation following
the structural modeling strategy of \citet{wu2020building}.
Three-dimensional geological models are generated by randomly
combining folding, faulting, and stratigraphic deformation.
Fold geometry is varied through changes in wavelength,
amplitude, and asymmetry. Fault systems include variations
in orientation, throw, spatial distribution, and assemblage
pattern. The resulting models provide both fault labels and
continuous RGT fields. For karst detection, we use the
same folded and faulted structural models but additionally
incorporate seismic reflection signatures associated with
karst caves of varying sizes. The corresponding cave regions
are used to generate supervision labels.

Synthetic seismic volumes are generated from these geological
models using randomized reflection properties and wavelets.
Two-dimensional sections are then sampled to form 440,000
training examples for fault detection, karst detection, and RGT estimation.
During training, the synthetic data
are further perturbed using additive noise, Gaussian
blurring, local blurring, and variations in wavelet dominant
frequency. These augmentations broaden the range of frequency
content, signal-to-noise ratio, and local imaging quality
represented by the synthetic training data.

A six-layer U-Net containing 24.23 million parameters is
used as the task-specific baseline across the three
interpretation tasks. It is initialized randomly and trained
using the corresponding synthetic training samples and
task-specific supervision.

For fault and karst detection, the SeisDFE encoder is kept fully frozen
and coupled with a DPT decoder \citep{ranftl2021vision}.
The training loss is
\begin{equation}
\mathcal{L}_{\mathrm{seg}}
=
0.7\mathcal{L}_{\mathrm{Dice}}
+
0.3\mathcal{L}_{\mathrm{BCE}},
\end{equation}
where the Dice term reduces the effect of foreground-background
imbalance and binary cross entropy provides pixel-wise
supervision.

RGT estimation requires continuous predictions that preserve
both local reflection geometry and global stratigraphic order.
We introduce LoRA modules into the last
30 Transformer blocks of SeisDFE, while a DPT decoder is
trained jointly with the LoRA parameters. The remaining encoder
parameters are unchanged. The RGT loss is
\begin{equation}
\mathcal{L}_{\mathrm{RGT}}
=
0.5\mathcal{L}_{1}
+
0.5\mathcal{L}_{\mathrm{SSIM}},
\end{equation}
where $\mathcal{L}_{1}$ constrains the predicted RGT values
and $\mathcal{L}_{\mathrm{SSIM}}$ promotes structural consistency.

The trained models are applied directly to unseen field
seismic sections. Because complete pixel-level labels are
unavailable for these field surveys, the synthetic-to-field
experiments are evaluated qualitatively by comparing
structural continuity and transfer behavior.

\begin{figure}[t]
\centering
\includegraphics[width=0.99\textwidth]{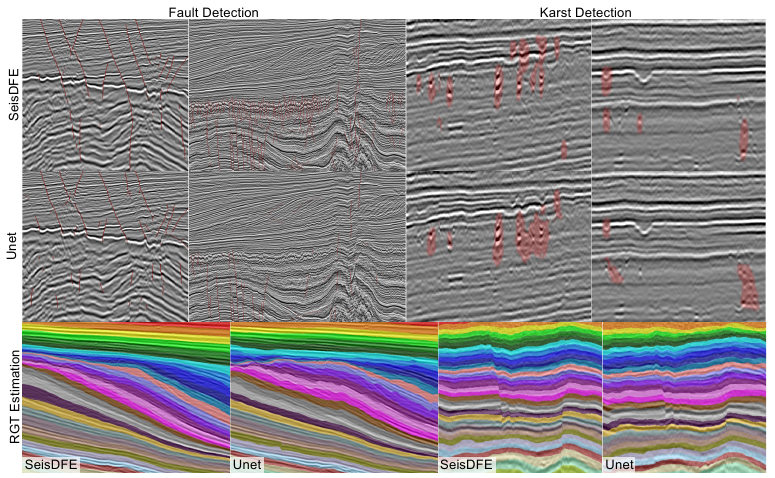}
\caption{Synthetic-to-field transfer on field seismic data. 
The upper two rows show two examples each of fault detection
(left) and karst cave detection (right), with predictions
obtained using SeisDFE and U-Net in the first and second rows,
respectively. The bottom row shows two RGT estimation examples,
each comparing SeisDFE (left) and U-Net (right).}
\label{fig4}
\end{figure}

For fault detection, both SeisDFE and U-Net recover
most major faults in the two field examples shown in the 
upper-left panels of Figure~\ref{fig4}.
Differences become more apparent for weak, short, or closely
spaced faults. SeisDFE produces more continuous responses in
several of these regions and shows fewer isolated detections
in structurally complex backgrounds. The result suggests that
the pretrained representation provides useful structural
information beyond that learned from synthetic fault labels
alone.

For karst detection, the models are trained using
synthetic seismic data containing cave-related reflection
signatures of varying sizes. Both models identify potential
cave-related anomalies in the two field examples
shown in the upper-right panels of Figure~\ref{fig4}. 
SeisDFE produces more spatially localized
responses in several regions, whereas U-Net tends to generate
broader detections in some areas. These qualitative differences
suggest that the pretrained representation can also support
the identification of localized geological anomalies.

For RGT estimation, both models
recover the overall stratigraphic trends in the two field
examples shown in the bottom row of
Figure~\ref{fig4}. SeisDFE retains finer variations
around folds, changes in dip, and faulted strata, while the
U-Net predictions are generally smoother. The pretrained
representation therefore appears to preserve more local
stratigraphic detail during synthetic-to-field transfer.

\begin{figure}[t]
\centering
\includegraphics[width=0.99\textwidth]{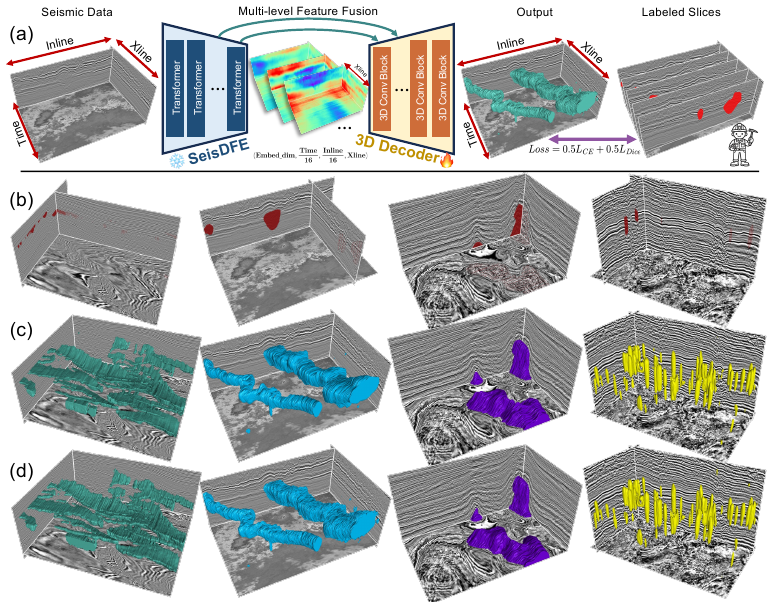}
\caption{Sparse 3D geobody detection. 
(a) Overview of the 3D interpretation framework, in which
multi-level features from neighboring seismic sections are
combined and processed by a 3D convolutional decoder under
sparse slice supervision.
(b) Sparse manual interpretations used for supervision.
(c) Predictions obtained with a 2D PUP decoder.
(d) Predictions obtained with the 3D convolutional decoder.
The four columns in (b)--(d), from left to right, show channel
detection on the Parihaka and Romney datasets, salt dome
detection on the F3 dataset, and karst detection
on a field seismic dataset.}
\label{fig5}
\end{figure}

Together, these examples show how synthetic supervision can be combined 
with a pretrained seismic representation
to support different interpretation tasks on unseen field
data. Most seismic surveys and interpretation targets,
however, are inherently three-dimensional, whereas SeisDFE
is pretrained on 2D seismic sections. We therefore further
test whether its 2D representations can be extended to
volumetric interpretation using only sparse slice annotations.

\subsection{Sparse 3D Interpretation}
\label{sec:sparse3d}

We evaluate two strategies for extending the frozen 2D
SeisDFE representation to 3D geobody interpretation
from sparse slice annotations. We use four datasets: 
channel interpretation on Parihaka and Romney, salt dome interpretation on F3,
and karst interpretation on a field seismic survey.

In the first strategy, each seismic section is encoded independently by SeisDFE
and interpreted using a 2D PUP decoder. Only annotated sections contribute to
training. At inference time, all sections are processed and reassembled into the
original 3D volume.

The second strategy introduces cross-slice information
during decoding, as illustrated in Figure~\ref{fig5}a.
A local window of $K$ neighboring seismic sections is
encoded independently by the frozen SeisDFE.
Multi-level features extracted from different Transformer
blocks are fused for each section before cross-slice
decoding. For section $k$, the resulting dense feature
map is denoted by
\begin{equation}
\mathbf{F}_{k}
\in
\mathbb{R}^{H'\times W'\times C},
\qquad
k=1,\ldots,K.
\end{equation}
The features are stacked according to their original spatial order:
\begin{equation}
\mathbf{F}_{3D}
=
\mathrm{Stack}
\left(
\mathbf{F}_{1},
\mathbf{F}_{2},
\ldots,
\mathbf{F}_{K}
\right)
\in
\mathbb{R}^{K\times H'\times W'\times C}.
\end{equation}
A lightweight 3D convolutional decoder then models both within-slice and
cross-slice relationships. We refer to this configuration as pseudo-3D because
the foundation encoder remains two-dimensional and 3D context is introduced only
at the decoding stage.

Supervision is available only on a subset of the volume. Unannotated voxels are
therefore excluded from the training loss.
Let $M_i\in\{0,1\}$ denote the annotation mask for voxel $i$, where
$M_i=1$ indicates an annotated voxel. The sparse supervision loss is
\begin{equation}
\mathcal{L}_{\mathrm{sparse}}
=
\frac{
\sum_i M_i\,\ell(\hat{y}_i,y_i)
}{
\sum_i M_i+\epsilon
},
\end{equation}
where $\hat{y}_i$ and $y_i$ denote the prediction and interpretation label,
respectively, and $\ell(\cdot)$ denotes the supervision loss in a general form.
In practice, we combine BCE and Dice losses with equal
weights of 0.5 for each annotated slice. The expression above emphasizes
that only annotated regions contribute to training, with BCE computed
pixel-wise and Dice evaluated over each annotated slice.
The BCE loss provides stable pixel-wise supervision, while the Dice loss
mitigates foreground-background imbalance, helping stabilize training.

During inference, overlapping 3D windows are processed across the full seismic
volume. Predictions in overlapping regions are fused to reduce window-boundary
artifacts and improve spatial stability.
The seismic volumes, sparse annotations, and predicted geobodies
in Figure~\ref{fig5} were visualized using CIGVis
\citep{li2025cigvis}.

Both approaches recover the main geobody distributions
across the four examples (Figure~\ref{fig5}c,d).
The 2D PUP decoder captures the overall geobody geometry,
but independent section-wise predictions can introduce
local fluctuations and fragmented boundaries.
The 3D convolutional decoder preserves the main geobody
shapes while providing more coherent cross-slice
continuity in several regions, particularly along
the elongated channel structures. The salt dome and
karst cave examples further demonstrate the applicability
of the same pretrained representation to detecting
geobodies with different spatial geometries.

Because complete voxel-level annotations are unavailable,
these experiments mainly assess the spatial coherence
of the detected geobodies. The results demonstrate that frozen 2D seismic
foundation features can support different types of
3D geobody detection through lightweight cross-slice
decoding, providing a practical alternative to training
a dedicated 3D foundation encoder when volumetric
annotations and computational resources are limited.


\section{Discussion}
\label{sec:discussion}
The benefits of seismic-domain pretraining vary across geological
targets. SeisDFE achieves substantial improvements in seismic facies
classification, salt detection, and RGT estimation. For fault detection,
it outperforms the other seismic foundation models but remains slightly
below its natural-image initialization, DINOv3-7B. This difference
reflects the spatial characteristics of the targets. Seismic facies,
salt bodies, and stratigraphic structures are expressed as spatially
extended reflection patterns, whereas faults are narrow, localized
discontinuities. The learned representation is therefore most effective
at capturing regional geological patterns and stratigraphic
relationships. The larger reduction of SeisDFE-Large on faults further
indicates that fine-scale structural information is more sensitive to
model compression. Structure-aware sampling, boundary-sensitive
augmentation, and objectives that emphasize structural discontinuities
are promising directions for strengthening fault-related representations.

Pretraining data diversity offers a direct route to further improvement.
Because neighboring sections and overlapping patches are strongly
correlated, the effective diversity of the corpus is determined more by
the number of distinct surveys and structural styles than by the number
of training images. Different geological settings contribute variations
in reflection geometry, structural complexity, and seismic character,
which are central to learning generalizable representations. Expanding
the coverage of complex structures, including faults, folds,
unconformities, and geobodies, is therefore a natural next step. Data
curation strategies such as feature-based retrieval, deduplication,
clustering, and distribution balancing
\citep{oquab2023dinov2,simeoni2025dinov3} can guide this expansion by
organizing the corpus according to geological structure, survey
provenance, and data quality.

The synthetic-to-field and sparse 3D experiments show how the pretrained
representation can be combined with readily available supervision.
In synthetic-to-field transfer, only the decoder, together with
low-rank adapters for RGT estimation, is trained on synthetic data,
whereas the encoder features were learned from field surveys. The
decoder therefore operates in a feature space already shaped by field
data, which narrows the domain gap that a randomly initialized U-Net
must bridge on its own. The more continuous fault responses, more
localized karst detections, and finer stratigraphic detail obtained
with SeisDFE (Figure~\ref{fig4}) reflect this advantage. In sparse 3D
interpretation, stacking features from neighboring sections and
decoding them with a lightweight 3D network improves cross-slice
continuity and produces more coherent geobody geometries
(Figure~\ref{fig5}c,d) without requiring a dedicated 3D foundation
encoder. Pretraining on sections from multiple orientations and
incorporating unlabeled data from target surveys could further improve
cross-survey transfer and three-dimensional structural consistency.

The present evaluation has several limitations.
First, complete reference labels are unavailable for the
field fault, karst, and RGT experiments (Figure~\ref{fig4}),
so these experiments primarily provide qualitative evidence
of synthetic-to-field transfer.
Similarly, the 3D interpretation experiments use only
sparsely annotated slices (Figure~\ref{fig5}b), limiting
the quantitative assessment of volumetric prediction accuracy.
More comprehensive evaluation requires denser reference
interpretations and independent geological validation.
In addition, the compared foundation models differ in
architecture, parameter scale, pretraining data, and
adaptation strategies. These differences complicate the
attribution of performance gains to individual pretraining
components.
Future work could incorporate standardized multi-survey
benchmarks, blind cross-survey evaluation, and uncertainty
analysis to further assess the generalization and
reliability of seismic foundation models.

\section{Conclusion}
\label{sec:conclusion}
We developed SeisDFE, a 7B-parameter seismic foundation model 
that learns reusable dense representations from 5.34 million unlabeled 
images drawn from more than 500 seismic volumes. 
Its three-stage pretraining yields features that are both discriminative 
and spatially coherent, and these properties are largely retained in the 
distilled SeisDFE-Large. With the encoder frozen, SeisDFE outperforms 
existing seismic foundation models across facies classification, salt detection, 
fault detection, and RGT estimation, with the largest gains when only a 
few annotated sections are available. Seismic-domain pretraining benefits 
spatially extended targets most, whereas fault detection depends more on spatial 
resolution and decoder design. The same frozen features also support 
synthetic-to-field transfer and sparse 3D geobody interpretation, positioning 
SeisDFE as a shared feature basis for data-efficient seismic interpretation.



\section*{Code availability}

The source code for SeisDFE pretraining, dense feature extraction,
model distillation, and downstream evaluation, together with training
configurations, is publicly available at
\url{https://github.com/ProgrammerZXG/SeisDFE}.

\section*{Competing interests}

The authors declare no competing interests.

\section*{Funding}

This work was supported by the National Natural Science
Foundation of China (NSFC) under Grant No. 42625404.


\bibliographystyle{plainnat}
\bibliography{ref}

@article{harsuko2022storseismic,
  title={StorSeismic: A new paradigm in deep learning for seismic processing},
  author={Harsuko, Randy and Alkhalifah, Tariq},
  journal={arXiv preprint arXiv:2205.00222},
  year={2022}
}

@article{bao2021beit,
  title={Beit: Bert pre-training of image transformers},
  author={Bao, Hangbo and Dong, Li and Piao, Songhao and Wei, Furu},
  journal={arXiv preprint arXiv:2106.08254},
  year={2021}
}

@article{gao2021channelseg3d,
  title={ChannelSeg3D: Channel simulation and deep learning for channel interpretation in 3D seismic images},
  author={Gao, Hang and Wu, Xinming and Liu, Guofeng},
  journal={Geophysics},
  volume={86},
  number={4},
  pages={IM73--IM83},
  year={2021},
  publisher={Society of Exploration Geophysicists}
}

@article{kaur2023deep,
  title={A deep learning framework for seismic facies classification},
  author={Kaur, Harpreet and Pham, Nam and Fomel, Sergey and Geng, Zhicheng and Decker, Luke and Gremillion, Ben and Jervis, Michael and Abma, Ray and Gao, Shuang},
  journal={Interpretation},
  volume={11},
  number={1},
  pages={T107--T116},
  year={2023},
  publisher={Society of Exploration Geophysicists and American Association of Petroleum~…}
}

@article{zhang2021automatic,
  title={Automatic seismic facies interpretation using supervised deep learning},
  author={Zhang, Haoran and Chen, Tiansheng and Liu, Yang and Zhang, Yuxi and Liu, Jiong},
  journal={Geophysics},
  volume={86},
  number={1},
  pages={IM15--IM33},
  year={2021},
  publisher={Society of Exploration Geophysicists}
}

@article{khosro2024machine,
  title={Machine learning for seismic exploration: Where are we and how far are we from the holy grail?},
  author={Khosro Anjom, Farbod and Vaccarino, Francesco and Socco, Laura Valentina},
  journal={Geophysics},
  volume={89},
  number={1},
  pages={WA157--WA178},
  year={2024},
  publisher={Society of Exploration Geophysicists}
}

@article{wysoczanska2024clip,
  author    = {Wysocza{\'n}ska, Monika and
               Ramamonjisoa, Micha{\"e}l and
               Trzci{\'n}ski, Tomasz and
               Sim{\'e}oni, Oriane},
  title     = {{CLIP-DIY}: {CLIP} Dense Inference Yields Open-Vocabulary
               Semantic Segmentation For-Free},
  journal   = {Proceedings of the IEEE/CVF Winter Conference on Applications of Computer Vision},
  year      = {2024},
  pages     = {1403--1413}
}

@article{alaudah2019machine,
  title={A machine-learning benchmark for facies classification},
  author={Alaudah, Yazeed and Micha{\l}owicz, Patrycja and Alfarraj, Motaz and AlRegib, Ghassan},
  journal={Interpretation},
  volume={7},
  number={3},
  pages={SE175--SE187},
  year={2019},
  publisher={Society of Exploration Geophysicists and American Association of Petroleum~…}
}

@article{geng2020deep,
  title={Deep learning for relative geologic time and seismic horizons},
  author={Geng, Zhicheng and Wu, Xinming and Shi, Yunzhi and Fomel, Sergey},
  journal={Geophysics},
  volume={85},
  number={4},
  pages={WA87--WA100},
  year={2020},
  publisher={Society of Exploration Geophysicists}
}

@article{wu2023sensing,
  title={Sensing prior constraints in deep neural networks for solving exploration geophysical problems},
  author={Wu, Xinming and Ma, Jianwei and Si, Xu and Bi, Zhengfa and Yang, Jiarun and Gao, Hui and Xie, Dongzi and Guo, Zhixiang and Zhang, Jie},
  journal={Proceedings of the National Academy of Sciences},
  volume={120},
  number={23},
  pages={e2219573120},
  year={2023},
  publisher={National Acad Sciences}
}

@article{wu2020building,
  title={Building realistic structure models to train convolutional neural networks for seismic structural interpretation},
  author={Wu, Xinming and Geng, Zhicheng and Shi, Yunzhi and Pham, Nam and Fomel, Sergey and Caumon, Guillaume},
  journal={Geophysics},
  volume={85},
  number={4},
  pages={WA27--WA39},
  year={2020},
  publisher={Society of Exploration Geophysicists}
}

@article{he2022masked,
  title={Masked autoencoders are scalable vision learners},
  author={He, Kaiming and Chen, Xinlei and Xie, Saining and Li, Yanghao and Doll{\'a}r, Piotr and Girshick, Ross},
  journal={Proceedings of the IEEE/CVF conference on computer vision and pattern recognition},
  pages={16000--16009},
  year={2022}
}

@article{oquab2023dinov2,
  title={Dinov2: Learning robust visual features without supervision},
  author={Oquab, Maxime and Darcet, Timoth{\'e}e and Moutakanni, Th{\'e}o and Vo, Huy and Szafraniec, Marc and Khalidov, Vasil and Fernandez, Pierre and Haziza, Daniel and Massa, Francisco and El-Nouby, Alaaeldin and others},
  journal={arXiv preprint arXiv:2304.07193},
  year={2023}
}

@article{archibong2025seiscode,
  title={SeisCoDE: 3D Seismic Interpretation Foundation Model with Contrastive Self-Distillation Learning},
  author={Archibong, Goodluck and Koeshidayatullah, Ardiansyah and Waheed, Umair and Li, Weichang and Harishidayat, Dicky and Alfarraj, Motaz},
  journal={arXiv preprint arXiv:2505.20518},
  year={2025}
}

@article{caron2021emerging,
  author    = {Caron, Mathilde and Touvron, Hugo and Misra, Ishan and
               J{\'e}gou, Herv{\'e} and Mairal, Julien and
               Bojanowski, Piotr and Joulin, Armand},
  title     = {Emerging Properties in Self-Supervised Vision Transformers},
  journal = {Proceedings of the IEEE/CVF International Conference on Computer Vision},
  year      = {2021},
  pages     = {9650--9660},
  doi       = {10.1109/ICCV48922.2021.00951}
}

@article{zhou2021ibot,
  title={ibot: Image bert pre-training with online tokenizer},
  author={Zhou, Jinghao and Wei, Chen and Wang, Huiyu and Shen, Wei and Xie, Cihang and Yuille, Alan and Kong, Tao},
  journal={arXiv preprint arXiv:2111.07832},
  year={2021}
}

@article{hu2021lora,
  title={Lora: Low-rank adaptation of large language models},
  author={Hu, Edward J and Shen, Yelong and Wallis, Phillip and Allen-Zhu, Zeyuan and Li, Yuanzhi and Wang, Shean and Wang, Lu and Chen, Weizhu},
  journal={arXiv preprint arXiv:2106.09685},
  year={2021}
}

@article{zheng2021rethinking,
  author    = {Zheng, Sixiao and Lu, Jiachen and Zhao, Hengshuang and
               Zhu, Xiatian and Luo, Zekun and Wang, Yabiao and
               Fu, Yanwei and Feng, Jianfeng and Xiang, Tao and
               Torr, Philip H. S. and Zhang, Li},
  title     = {Rethinking Semantic Segmentation from a Sequence-to-Sequence
               Perspective with Transformers},
  journal   = {Proceedings of the IEEE/CVF Conference on Computer Vision and Pattern Recognition},
  year      = {2021},
  pages     = {6881--6890},
  doi       = {10.1109/CVPR46437.2021.00681}
}

@article{ranftl2021vision,
  author    = {Ranftl, Ren{\'e} and Bochkovskiy, Alexey and Koltun, Vladlen},
  title     = {Vision Transformers for Dense Prediction},
  journal   = {Proceedings of the IEEE/CVF International Conference on Computer Vision},
  year      = {2021},
  pages     = {12179--12188},
  doi       = {10.1109/ICCV48922.2021.01196}
}

@article{yu2021deep,
  title={Deep learning for geophysics: Current and future trends},
  author={Yu, Siwei and Ma, Jianwei},
  journal={Reviews of Geophysics},
  volume={59},
  number={3},
  pages={e2021RG000742},
  year={2021},
  publisher={Wiley Online Library}
}

@article{zhou2023foundation,
  title={A foundation model for generalizable disease detection from retinal images},
  author={{Zhou}, Yukun and Chia, Mark A and Wagner, Siegfried K and Ayhan, Murat S and Williamson, Dominic J and Struyven, Robbert R and Liu, Timing and Xu, Moucheng and Lozano, Mateo G and Woodward-Court, Peter and others},
  journal={Nature},
  volume={622},
  number={7981},
  pages={156--163},
  year={2023},
  publisher={Nature Publishing Group UK London}
}

@article{wu2019faultseg3d,
  title={FaultSeg3D: Using synthetic data sets to train an end-to-end convolutional neural network for 3D seismic fault segmentation},
  author={Wu, Xinming and Liang, Luming and Shi, Yunzhi and Fomel, Sergey},
  journal={Geophysics},
  volume={84},
  number={3},
  pages={IM35--IM45},
  year={2019},
  publisher={Society of Exploration Geophysicists}
}

@article{roncoroni2022efficient,
  title={Efficient extraction of seismic reflection with Deep Learning},
  author={Roncoroni, Giacomo and Forte, Emanuele and Bortolussi, Luca and Pipan, Michele},
  journal={Computers \& Geosciences},
  volume={166},
  pages={105190},
  year={2022},
  publisher={Elsevier}
}

@article{schafer2024overcoming,
  title={Overcoming data scarcity in biomedical imaging with a foundational multi-task model},
  author={Sch{\"a}fer, Raphael and Nicke, Till and H{\"o}fener, Henning and Lange, Annkristin and Merhof, Dorit and Feuerhake, Friedrich and Schulz, Volkmar and Lotz, Johannes and Kiessling, Fabian},
  journal={Nature Computational Science},
  volume={4},
  number={7},
  pages={495--509},
  year={2024},
  publisher={Nature Publishing Group US New York}
}

@article{yan20253d,
  title={3D karst cave recognition using TransUnet with dual attention mechanisms in seismic images},
  author={Yan, Binpeng and Zhao, Jiaqi and Peng, Kuankuan and Qian, Longlong and Li, Mutian and Pan, Rui},
  journal={Geophysics},
  volume={90},
  number={5},
  pages={IM133--IM143},
  year={2025},
  publisher={Society of Exploration Geophysicists}
}

@article{di2019developing,
  title={Developing a seismic texture analysis neural network for machine-aided seismic pattern recognition and classification},
  author={Di, Haibin and Gao, Dengliang and AlRegib, Ghassan},
  journal={Geophysical Journal International},
  volume={218},
  number={2},
  pages={1262--1275},
  year={2019},
  publisher={Oxford University Press}
}

@article{chai2020using,
  title={Using a deep neural network and transfer learning to bridge scales for seismic phase picking},
  author={Chai, Chengping and Maceira, Monica and Santos-Villalobos, Hector J and Venkatakrishnan, Singanallur V and Schoenball, Martin and Zhu, Weiqiang and Beroza, Gregory C and Thurber, Clifford and EGS Collab Team},
  journal={Geophysical Research Letters},
  volume={47},
  number={16},
  pages={e2020GL088651},
  year={2020},
  publisher={Wiley Online Library}
}

@article{mousavi2022deep,
  title={Deep-learning seismology},
  author={{Mousavi}, S Mostafa and Beroza, Gregory C},
  journal={Science},
  volume={377},
  number={6607},
  pages={eabm4470},
  year={2022},
  publisher={American Association for the Advancement of Science}
}

@article{geng2020automated,
  title={Automated design of a convolutional neural network with multi-scale filters for cost-efficient seismic data classification},
  author={Geng, Zhi and Wang, Yanfei},
  journal={Nature communications},
  volume={11},
  number={1},
  pages={3311},
  year={2020},
  publisher={Nature Publishing Group UK London}
}

@article{gao2025geologically,
  title={A Geologically Informed and Data-Driven AI Workflow for Fully Seismic Stratigraphic Interpretation of Sedimentary Basin},
  author={Gao, Hui and Wu, Xinming and Ding, Xuesong},
  journal={IEEE Transactions on Geoscience and Remote Sensing},
  volume={63},
  pages={1--13},
  year={2025},
  publisher={IEEE}
}

@misc{tgs,
    author = {Addison Howard, Arvind Sharma and Ashleigh Lenamond, cenyen and Compu Ter, John Adamck and Mark McDonald, Sathiya and Sri Kainkaryam, Will Cukierski},
    title = {TGS Salt Identification Challenge},
    publisher = {Kaggle},
    year = {2018},
    url = {https://kaggle.com/competitions/tgs-salt-identification-challenge}
}

@misc{seam,
    author = {SEAM},
    title = {Facies Identification Challenge: 3D image interpretation by machine learning techniques},
    publisher = {AIcrowd},
    year = {2020},
    url = {https://www.aicrowd.com/challenges/seismic-facies-identification-challenge}
}

@article{guo2025cross,
  title={Cross-domain foundation model adaptation: Pioneering computer vision models for geophysical data analysis},
  author={Guo, Zhixiang and Wu, Xinming and Liang, Luming and Sheng, Hanlin and Chen, Nuo and Bi, Zhengfa},
  journal={Journal of Geophysical Research: Machine Learning and Computation},
  volume={2},
  number={1},
  pages={e2025JH000601},
  year={2025},
  publisher={Wiley Online Library}
}

@article{gao2026foundation,
  title={A foundation model enpowered by a multi-modal prompt engine for universal seismic geobody interpretation across surveys},
  author={Gao, Hang and Wu, Xinming and Liang, Luming and Sheng, Hanlin and Si, Xu and Gao, Hui and Li, Yaxing},
  journal={Information Fusion},
  volume={125},
  pages={103437},
  year={2026},
  publisher={Elsevier},
  doi={10.1016/j.inffus.2025.103437}
}

@article{liu2020seismic,
  title={Seismic facies classification using supervised convolutional neural networks and semisupervised generative adversarial networks},
  author={Liu, Mingliang and Jervis, Michael and Li, Weichang and Nivlet, Philippe},
  journal={Geophysics},
  volume={85},
  number={4},
  pages={O47--O58},
  year={2020},
  publisher={Society of Exploration Geophysicists}
}

@article{dou2024faultssl,
  title={FaultSSL: Seismic fault detection via semisupervised learning},
  author={Dou, Yimin and Li, Kewen and Dong, Minghui and Xiao, Yuan},
  journal={Geophysics},
  volume={89},
  number={3},
  pages={M79--M91},
  year={2024},
  publisher={Society of Exploration Geophysicists}
}

@article{gao2022automatic,
  title={Automatic fault detection on seismic images using a multiscale attention convolutional neural network},
  author={Gao, Kai and Huang, Lianjie and Zheng, Yingcai and Lin, Rongrong and Hu, Hao and Cladohous, Trenton},
  journal={Geophysics},
  volume={87},
  number={1},
  pages={N13--N29},
  year={2022},
  publisher={Society of Exploration Geophysicists}
}

@article{wu2019multitask,
  title={Multitask learning for local seismic image processing: Fault detection, structure-oriented smoothing with edge-preserving, and seismic normal estimation by using a single convolutional neural network},
  author={Wu, Xinming and Liang, Luming and Shi, Yunzhi and Geng, Zhicheng and Fomel, Sergey},
  journal={Geophysical Journal International},
  volume={219},
  number={3},
  pages={2097--2109},
  year={2019},
  publisher={Oxford University Press}
}

@article{xiong2018seismic,
  title={Seismic fault detection with convolutional neural network},
  author={Xiong, Wei and Ji, Xu and Ma, Yue and Wang, Yuxiang and AlBinHassan, Nasher M and Ali, Mustafa N and Luo, Yi},
  journal={Geophysics},
  volume={83},
  number={5},
  pages={O97--O103},
  year={2018},
  publisher={Society of Exploration Geophysicists}
}

@article{wen2025semantic,
  title={Semantic concentration for self-supervised dense representations learning},
  author={Wen, Peisong and Xu, Qianqian and Dai, Siran and Cong, Runmin and Huang, Qingming},
  journal={IEEE Transactions on Pattern Analysis and Machine Intelligence},
  year={2025},
  publisher={IEEE}
}

@article{bardes2022vicregl,
  title={Vicregl: Self-supervised learning of local visual features},
  author={Bardes, Adrien and Ponce, Jean and LeCun, Yann},
  journal={Advances in neural information processing systems},
  volume={35},
  pages={8799--8810},
  year={2022}
}

@article{ordonez2026ncs,
  title={The NCS-Model: A seismic foundation model trained on the Norwegian repository of public data},
  author={Ordonez, Alba and Forgaard, Theodor Johannes Line and Wade, David and Bugge, Aina Juell and Nese, Hakon and Waldeland, Anders Ueland},
  journal={arXiv preprint arXiv:2603.23211},
  year={2026}
}

@misc{dong2025lightweightedfoundationmodelseismic,
      title={Light-weighted foundation model for seismic data processing based on representative and non-redundant pre-training dataset}, 
      author={Xintong Dong and Wenshuo Yu and Jun Lin and Zhenbo Guo and Hongzhou Wang and Jianhao Yang},
      year={2025},
      eprint={2503.10092},
      archivePrefix={arXiv},
      primaryClass={physics.geo-ph},
      url={https://arxiv.org/abs/2503.10092}, 
}

@misc{cheng2025generativefoundationmodelallinone,
      title={A generative foundation model for an all-in-one seismic processing framework}, 
      author={Shijun Cheng and Randy Harsuko and Tariq Alkhalifah},
      year={2025},
      eprint={2502.01111},
      archivePrefix={arXiv},
      primaryClass={physics.geo-ph},
      url={https://arxiv.org/abs/2502.01111}, 
}

@article{simeoni2025dinov3,
  title={{DINOv3}},
  author={Sim{\'e}oni, Oriane and Vo, Huy V. and Seitzer, Maximilian and Baldassarre, Federico and Oquab, Maxime and Jose, Cijo and Khalidov, Vasil and Szafraniec, Marc and Yi, Seungeun and Ramamonjisoa, Micha{\"e}l and Massa, Francisco and Haziza, Daniel and Wehrstedt, Luca and Wang, Jianyuan and Darcet, Timoth{\'e}e and Moutakanni, Th{\'e}o and Sentana, Leonel and Roberts, Claire and Vedaldi, Andrea and Tolan, Jamie and Brandt, John and Couprie, Camille and Mairal, Julien and J{\'e}gou, Herv{\'e} and Labatut, Patrick and Bojanowski, Piotr},
  journal={arXiv preprint arXiv:2508.10104},
  year={2025}
}

@article{su2024roformer,
  title={{RoFormer}: Enhanced transformer with rotary position embedding},
  author={Su, Jianlin and Ahmed, Murtadha and Lu, Yu and Pan, Shengfeng and Bo, Wen and Liu, Yunfeng},
  journal={Neurocomputing},
  volume={568},
  pages={127063},
  year={2024}
}

@article{sheng2025seismic,
  author={Sheng, Hanlin and Wu, Xinming and Si, Xu and Li, Jintao and Zhang, Sibo and Duan, Xudong},
  title={Seismic foundation model: A next generation deep-learning model in geophysics},
  journal={Geophysics},
  volume={90},
  number={2},
  pages={IM59--IM79},
  year={2025},
  doi={10.1190/geo2024-0262.1}
}

@article{li2025cigvis,
  title = {{CIGVis}: An open-source Python tool for the real-time interactive visualization of multidimensional geophysical data},
  author = {Li, Jintao and Shi, Yunzhi and Wu, Xinming},
  journal = {Geophysics},
  year = {2025},
  volume = {90},
  number = {1},
  pages = {F1--F10},
  doi = {10.1190/geo2024-0041.1},
  publisher = {Society of Exploration Geophysicists}
}

\end{document}